\documentclass[review]{elsarticle}

\usepackage{url}
\usepackage{lineno,hyperref}
\usepackage{tikz}
\def\checkmark{\tikz\fill[scale=0.4](0,.35) -- (.25,0) -- (1,.7) -- (.25,.15) -- cycle;} 
\modulolinenumbers[5]

\journal{Artificial Intelligence in Medicine}

\begin{document}

\begin{frontmatter}

\title{Deep Learning Based Decision Support for Medicine - A Case Study on Skin Cancer Diagnosis}
%\tnotetext[mytitlenote]{Fully documented templates are available in the elsarticle package on \href{http://www.ctan.org/tex-archive/macros/latex/contrib/elsarticle}{CTAN}.}

%% Group authors per affiliation:
\author[1,2]{Adriano Lucieri\corref{cor1}%
%\fnref{fn1}
}
\ead{adriano.lucieri@dfki.de}

\author[1,2]{Andreas Dengel}
\ead{andreas.dengel@dfki.de}

\author[1]{Sheraz Ahmed}
\ead{sheraz.ahmed@dfki.de}

\cortext[cor1]{Corresponding author}

\address[1]{Deutsches Forschungszentrum für Künstliche Intelligenz GmbH (DFKI), Smart Data \& Knowledge Services (SDS), Trippstadter Straße 122, 67663 Kaiserslautern, Germany}
\address[2]{TU Kaiserslautern, Erwin-Schrödinger-Straße 52, 67663 Kaiserslautern, Germany}

\begin{abstract}
Early detection of skin cancers like melanoma is crucial to ensure high chances of survival for patients. Clinical application of Deep Learning (DL)-based Decision Support Systems (DSS) for skin cancer screening has the potential to improve the quality of patient care. The majority of work in the medical AI community focuses on a diagnosis setting that is mainly relevant for autonomous operation. Practical decision support should, however, go beyond plain diagnosis and provide explanations. 
%Moreover, the integration of data-driven DSS raises questions of liability that   transparency and decision explanation. 
%One way to address this issue is to provide explanations. 
This paper provides an overview of works towards explainable, DL-based decision support in medical applications with the example of skin cancer diagnosis from clinical, dermoscopic and histopathologic images. 
%Summaries are provided for works addressing the classification of clinical, Dermoscopic and histological image data. 
Analysis reveals that comparably little attention is payed to the explanation of histopathologic skin images and that current work is dominated by visual relevance maps as well as dermoscopic feature identification.
%The practicality of the reviewed work is critically assessed and missing traits of current approaches are identified. 
%Novel paths of decision explanation are discussed, potentially increasing explanation diversity and therefore stakeholder trust. 
%Rethinking of current explanation attempts is a crucial step towards successful deployment of DL-based classifiers in general clinical practice.
We conclude that future work should focus on meeting the stakeholder's cognitive concepts, providing exhaustive explanations that combine global and local approaches and leverage diverse modalities. Moreover, the possibility to intervene and guide models in case of misbehaviour is identified as a major step towards successful deployment of AI as DL-based DSS and beyond.
%The final discussion points out future directions towards the deployment of explainable DL-based classifiers in clinical practice.
\end{abstract}


\end{frontmatter}

%\linenumbers

% Introduction : "Performance of a dermoscopy-based computer vision system for the diagnosis of pigmented skin lesions compared with visual evaluation by experienced dermatologists" ODER
% "Computerized analysis of pigmented skin lesions: A review" -- früh schon digital versucht was in derma zu machen

% "Do physicians value decision support? A look at the effect of decision support systems on physician opinion" --> Intorduction as well --> explanations would increase. Vielleicht doch nicht. Ist sehr alt

\section{Introduction}
In the last decade, Deep Learning (DL) and particularly the Convolutional Neural Network (CNN) has demonstrated its exceptional ability to efficiently solve a variety of image-based tasks in medical diagnosis 
%of diseases~\cite{esteva2017dermatologist,qiu2020development,brinker2019deep,abramoff2018pivotal}. 
%The sub-area of Artificial Intelligence (AI) has specially gained momentum with the ever increasing number of digital data that arises as a side product of our modern lives, and the affordability of computational resources. Not only private households continuously generate data, but also industries, authorities and the healthcare system. Structured digital data that has been evaluated and annotated by experts has an enormous value which is slowly being understood by more and more data owners. Apart from the development of new business cases that make use of behavioural user data and inventions that further increase high standards of living, there exist more noble ways of exploiting the power of structured data together with learning algorithms. Together with AI, our modern digital infrastructure is able to support people in various difficult situations. For instance, early warning systems for natural disasters like flooding~\cite{nevo2019ml}, wildfires~\cite{sayad2019predictive} and earthquakes~\cite{mignan2020neural} can save many lives and mitigate their consequences. The AI-based outbreak risk software BlueDot was among the first to detect the recent COVID-19 outbreak before 2020~\cite{bogoch2020potential}. 
%Similarly, DL showed to be effective in diagnosing a variety of diseases in patients like 
such as the detection of skin cancer~\cite{brinker2019deep}, Alzheimer's~\cite{qiu2020development} and Diabetic Retinopathy (DR)~\cite{abramoff2018pivotal}. With IDx-DR~\cite{abramoff2018pivotal} and 3DermSpot~\cite{PRNewswire2020}, Digital Diagnostics Inc. have launched the first ever autonomous Artificial Intelligence (AI) diagnostic systems, receiving breakthrough designation by the FDA\footnote{Food and Drug Administration}. Although having been approved in the United States (US) under close cooperation with the FDA~\cite{FDA2018}, the inner workings of DL-based black-box algorithms still raise numerous questions regarding their decision-making processes~\cite{rudin2019stop} and unexpected malfunctions~\cite{szegedy2014intriguing}. Bissoto et al.~\cite{bissoto2019constructing} explicitly demonstrate that \textit{Clever Hans} behaviour might often be mistaken with high diagnostic performance in the classification of skin cancer from  dermoscopic images. In contrast to the US, the strategy pursued by the European Union (EU) is heavily focusing on human oversight, transparency and building an ecosystem of trust~\cite{european2020white}. Specially the application of AI in the medical domain bears complex ethical and legal implications which demand the use of explainable and traceable solutions within such an ecosystem~\cite{schneeberger2020european}. Attempting to explain DL-based systems might also be the best way to validate a system's performance beyond quantitative metrics, preventing issues arising from effects as demonstrated in~\cite{bissoto2019constructing}. The development of explainable AI (xAI) and methods that explain existing AI algorithms is therefore a blooming research area.

%Often, patients, doctors and authorities restrain from applying opaque algorithms in such critical areas. This is further emphasized by the 2018 General Data Protection Regulation (GDPR)~\cite{GDPR2018} issued by the European Union (EU). It demands a right to explanation for automated decision affecting any individual. Moreover, attempting to explain DL-based systems might be the best way to validate their performance beyond quantitative metrics and learning from them. The development of explainable AI (xAI) and methods that explain existing AI algorithms is therefore a blooming research area.

Skin cancer is one of the most common types of cancer worldwide~\cite{khazaei2019global}. Melanoma is a particularly dangerous type of skin cancer. In the US, it accounts for only 1\% of skin cancers diagnosed, but resulting in the largest share of skin cancer related deaths~\cite{american20202cancer}. The manual diagnosis of skin lesions involves the regular examination of a patient's skin lesions by dermatologists that went through years of specialized training. The cancer is diagnosed by thoroughly analyzing skin lesions, applying rules of dermoscopic pattern recognition and algorithms like the ABCD-rule~\cite{nachbar1994abcd} or 7-point checklist~\cite{argenziano1998epiluminescence} that have been developed and described in medical literature throughout years of research. 
This routine process requires constant attention of the expert. Therefore, diagnostic performance is highly dependent on fatigue and emotional state of the physician.
%Apart from being tedious work that requires the attention of well-trained professionals, it has also been shown that the recognition of dermoscopic criteria is highly subjective and therefore doctor's decisions hard to reproduce~\cite{elmore2017pathologists}
This and the subjectivity of dermoscopic criteria recognition might be contributing to the fact that doctor's decisions are hard to reproduce~\cite{elmore2017pathologists}.
The augmentation of human doctors through AI-based Decision Support Systems (DSS) is highly desirable as it potentially increases reproducibility of results, speeding up tedious examinations, and therefore allowing more thorough treatment of the broader population. Fortunately, initiatives for public large-scale repositories of digital skin images such as the International Skin Imaging Collaboration (ISIC)\footnote{https://www.isic-archive.com/} drove an increasing interest of AI application in this domain.

This work provides an overview of current explainability methods for DLS applied to the problem of skin lesion classification. The critical assessment of current approaches ought to present the bigger picture of AI in dermatology to identify missing traits that impede the deployment of AI assistants in the medical domain. Section~\ref{sec:background} provides basic knowledge of skin lesion classification and common explainability methods for DL-based classifiers. Then, the reviewed work is summarized and classified into four non-exclusive groups describing the main explanation techniques in Section~\ref{sec:overview}. Section~\ref{sec:discussion} critically assesses the current state of explainable DL-based skin lesion classification. Moreover, four major objectives to be followed for clinically practical DLS in medicine are proposed, followed by concluding remarks in Section~\ref{sec:conclusion}.

% \section{Related Work}
% Holzinger et al.~\cite{holzinger2017we}
% Recently, Singh et al.~\cite{singh2020explainable} reviewed literature on explanation methods for DL-based classifiers applied to applications of MIA. Tjoa et al.~\cite{tjoa2019survey} review general literature of explainable AI, trying to discuss the road towards AI applications in medicine.
% Gilvary et al.~\cite{gilvary2019missing} identify stuff.. And He et al.~\cite{he2019practical} do some other stuff.

\section{Background}
\label{sec:background}

\subsection{Skin Cancer Diagnosis}
The classification of skin lesions is usually attempted by analysing one of the three most popular domains, i.e. clinical, dermoscopic and histopathologic imaging. \textbf{Clinical images} are digitized naked-eye observations of the skin and exhibit the lowest information value. \textbf{Dermoscopic images} are taken using a dermatoscope, which uses light and high magnification rates to visualize patterns present in upper layers of skin in more detail. \textbf{Histopathology} is the microscopic examination of cell tissue. Whole Slide Images (WSIs) are detailed images of enormous size (up to Gigapixels) that are generated by scanning glass slides of skin tissue. Histopathologic examination is regarded as the gold standard in skin lesion diagnosis, whereas the information content declines for dermoscopic and clinical images accordingly.

\subsection{XAI Taxonomy for Skin Cancer Diagnosis}
%The explanation of DL-based classifiers is a trending research topic. 
%Some of the various proposed methods relevant to the reviewed work will be introduced hereinafter. 
A variety of taxonomies for xAI methods exist in literature~\cite{arya2019one, arrieta2020explainable}. In contrast to existing categorizations for the broad application of xAI methods, we group the reviewed work in four main categories, i.e. \textit{Visual Relevance Localization}, \textit{Dermoscopic Feature Prediction \& Localization}, \textit{Similarity Retrieval} and \textit{Intervention}. This distinction is specifically selected to fit the dermatological use-case as well as being user-centric, focusing on the utility of the output rather than the producing mechanisms. An overview of the reviewed works in the four non-exclusive categories is presented in Fig.~\ref{fig:taxonomy}. 

% \begin{figure}
% \centering
% \includegraphics[width=\textwidth]{figures/Taxonomy.png}
% \caption{Overview of reviewed publications, grouped by the proposed taxonomy. Single papers can belong to multiple groups.}
% \label{fig:taxonomy}
% \end{figure}

\begin{figure}
    \centering
    \begin{tikzpicture}[]
%\draw[step=0.5cm,gray,very thin] (0,0) grid (11.5,6);
\node[align=center, anchor=center] at (6,6) (title) {Attempts on Explainable};
\node[below of=title, node distance=1.0em] {AI in Dermatology};

% Upper left row
\filldraw[fill=red!40!white, draw=black, opacity=0.3, rounded corners=0.3em] (0,2.25) rectangle (8.625-0.2, 5.25);
\node[align=center, anchor=center] at (4.3125-0.1,5) {\small Visual Relevance Localization};

\filldraw[fill=red!40!white, draw=black, opacity=0.3, rounded corners=0.3em] (0,3.75) rectangle (1.4375*2, 4.75);
\node[align=center, anchor=center] at (1.4375,4.65) (act) {\scriptsize Activation};
\node[below of=act, node distance=1.0em] (act_1) {\tiny Van Molle et al.~\cite{van2018visualizing}};
\node[below of=act_1, node distance=0.65em] (act_2) {\tiny Barata et al.~\cite{barata2020explainable}};

\filldraw[fill=red!40!white, draw=black, opacity=0.3, rounded corners=0.3em] (1.4375*2,2.25) rectangle (1.4375*6-0.2, 4.75);
\node[align=center, anchor=center] at (5.75,4.65) (attr) {\scriptsize Attribution};
\node[align=center, anchor=center] at (4.3125+0.2,4.65) (attr_l) {\scriptsize};
\node[below of=attr_l, node distance=1.0em] (attr_1) {\tiny Cruz-Roa et al.~\cite{cruz2013deep}};
\node[below of=attr_1, node distance=0.65em] (attr_2) {\tiny Cruz-Roa et al.~\cite{cruz2014automatic}};
\node[below of=attr_2, node distance=0.65em] (attr_3) {\tiny Radhakrishnan et al.~\cite{radhakrishnan2017patchnet}};
\node[below of=attr_3, node distance=0.65em] (attr_4) {\tiny Esteva et al.~\cite{esteva2017dermatologist}};
\node[below of=attr_4, node distance=0.65em] (attr_5) {\tiny Ge et al.~\cite{ge2017skin}};
\node[below of=attr_5, node distance=0.65em] (attr_6) {\tiny Jia et al.~\cite{jia2017skin}};
\node[below of=attr_6, node distance=0.65em] (attr_7) {\tiny Li et al.~\cite{li2018evidence}};
\node[below of=attr_7, node distance=0.65em] (attr_8) {\tiny Codella et al.~\cite{codella2018collaborative}};
\node[below of=attr_8, node distance=0.65em] (attr_9) {\tiny Thandiackal et al.~\cite{thandiackal2018structure}};
\node[align=center, anchor=center] at (7.1875-0.2,4.65) (attr_r) {\scriptsize};
\node[below of=attr_r, node distance=1.0em] (attr_10) {\tiny Yang et al.~\cite{yang2018classification}};
\node[below of=attr_10, node distance=0.65em] (attr_11) {\tiny Mishra et al.~\cite{mishra2019interpreting}};
\node[below of=attr_11, node distance=0.65em] (attr_12) {\tiny Xiang et al.~\cite{xiang2019towards}};
\node[below of=attr_12, node distance=0.65em] (attr_13) {\tiny Rieger et al.~\cite{rieger2019interpretations}};
\node[below of=attr_13, node distance=0.65em] (attr_14) {\tiny Young et al.~\cite{young2019deep}};
\node[below of=attr_14, node distance=0.65em] (attr_15) {\tiny Xie et al.~\cite{xie2019interpretable}};
\node[below of=attr_15, node distance=0.65em] (attr_16) {\tiny Miko\l{}ajczyk~\cite{mikolajczyk2020global}};
\node[below of=attr_16, node distance=0.65em] (attr_17) {\tiny Sonntag et al.~\cite{sonntag2020skincare}};
\node[below of=attr_17, node distance=0.65em] (attr_18) {\tiny Hägele et al.~\cite{hagele2020resolving}};

\filldraw[fill=red!40!white, draw=black, opacity=0.3, rounded corners=0.3em] (0,2.25) rectangle (1.4375*2, 3.75);
\node[align=center, anchor=center] at (1.4375,3.525) (att) {\scriptsize Attention};
\node[below of=att, node distance=1.0em] (att_1) {\tiny Jia et al.~\cite{jia2017skin}};
\node[below of=att_1, node distance=0.65em] (att_2) {\tiny Yan et al.~\cite{yan2019melanoma}};
\node[below of=att_2, node distance=0.65em] (att_3) {\tiny Zhang et al.~\cite{zhang2017mdnet}};
\node[below of=att_3, node distance=0.65em] (att_4) {\tiny Barata et al.~\cite{barata2020explainable}};

% Upper right row
\filldraw[fill=orange!40!white, draw=black, opacity=0.3, rounded corners=0.3em] (8.625-0.2,2.25) rectangle (11.5,5.25);
\node[align=center, anchor=center] at (10.0625, 5) (sim) {\small Similarity Retrieval};
\node[below of=sim, node distance=1.0em] (sim_1) {\tiny Kawahara et al.~\cite{kawahara2018seven}};
\node[below of=sim_1, node distance=0.65em] (sim_1) {\tiny Codella et al.~\cite{codella2018collaborative}};

% Bottom mid cell
\filldraw[fill=blue!40!white, draw=black, opacity=0.3, rounded corners=0.3em] (0.0,0) rectangle (11.5,2.25);
\node[align=center, anchor=center] at (6.075,1.9) (DF) {\small Dermoscopic Feature Prediction \& Localization};

\filldraw[fill=blue!40!white, draw=black, opacity=0.3, rounded corners=0.3em] (0.0,0) rectangle (5.75,1.6);
\node[align=center, anchor=center] at (2.875,1.4) (DFP) {\scriptsize Prediction};
\node[align=center, anchor=center] at (1.4375+0.2,1.4) (DFP_l) {\scriptsize};
\node[below of=DFP_l, node distance=1.0em] (DFP_1) {\tiny Kawahara et al.~\cite{kawahara2018fully}};
\node[below of=DFP_1, node distance=0.65em] (DFP_2) {\tiny Codella et al.~\cite{codella2018collaborative}};
\node[below of=DFP_2, node distance=0.65em] (DFP_3) {\tiny Veltmeijer et al.~\cite{veltmeijer2019integrating}};
\node[below of=DFP_3, node distance=0.65em] (DFP_4) {\tiny Murabayashi et al.~\cite{murabayashi2019towards}};
\node[align=center, anchor=center] at (4.3125-0.2,1.4) (DFP_r) {\scriptsize};
\node[below of=DFP_r, node distance=1.0em] (DFP_5) {\tiny Lucieri et al.~\cite{lucieri2020interpretability}};
\node[below of=DFP_5, node distance=0.65em] (DFP_6) {\tiny Coppola et al.~\cite{coppola2020interpreting}};
\node[below of=DFP_6, node distance=0.65em] (DFP_7) {\tiny Chen et al.~\cite{chen2020concept}};

\filldraw[fill=blue!40!white, draw=black, opacity=0.3, rounded corners=0.3em] (5.75,0) rectangle (11.5,1.6);
\node[align=center, anchor=center] at (8.625,1.4) (DFL) {\scriptsize Localization};
\node[align=center, anchor=center] at (7.1875+0.2,1.4) (DFL_l) {\scriptsize};
\node[below of=DFL_l, node distance=1.0em] (DFL_1) {\tiny Kawahara et al.~\cite{kawahara2018seven}};
\node[below of=DFL_1, node distance=0.65em] (DFL_2) {\tiny Kawahara et al.~\cite{kawahara2018fully}};
\node[below of=DFL_2, node distance=0.65em] (DFL_3) {\tiny Thandiackal et al.~\cite{thandiackal2018structure}};
\node[align=center, anchor=center] at (10.0625-0.2,1.4) (DFL_r) {\scriptsize};
\node[below of=DFL_r, node distance=1.0em] (DFL_4) {\tiny Zhang et al.~\cite{zhang2019biomarker}};
\node[below of=DFL_4, node distance=0.65em] (DFL_5) {\tiny Veltmeijer et al.~\cite{veltmeijer2019integrating}};
\node[below of=DFL_5, node distance=0.65em] (DFL_6) {\tiny Sonntag et al.~\cite{sonntag2020skincare}};

% Bottom right row
\filldraw[fill=green!40!white, draw=black, opacity=0.3, rounded corners=0.3em] (8.625-0.2,2.25) rectangle (11.5,3.75);
\node[align=center, anchor=center] at (10.0625,3.5) (inter) {\small Intervention};
\node[below of=inter, node distance=1.0em] (inter_1) {\tiny Yan et al.~\cite{yan2019melanoma}};
\node[below of=inter_1, node distance=0.65em] (inter_2) {\tiny Rieger et al.~\cite{rieger2019interpretations}};
\node[below of=inter_2, node distance=0.65em] (inter_3) {\tiny Miko\l{}ajczyk~\cite{mikolajczyk2020global}};
\node[below of=inter_3, node distance=0.65em] (inter_4) {\tiny Chen et al.~\cite{chen2020concept}};

\end{tikzpicture}
    \caption{Overview of reviewed publications, grouped by the proposed taxonomy. Single papers can belong to multiple groups.}
    \label{fig:taxonomy}
\end{figure}

\paragraph{Visual Relevance Localization} A popular approach towards explaining CNNs is the generation of visual maps that provide spatial information about the input's relevance to a prediction. 
%Those methods are hereinafter referred to as \textit{Visual Relevance Localization} methods. 
We further distinguish post-hoc, relevance-based methods that visualize network's activations (\textit{Activation} e.g.~\cite{zeiler2014visualizing,simonyan2013deep}) or trace the input's relevance to prediction (\textit{Attribution}, e.g. \cite{shrikumar2016not,zhou2016learning,selvaraju2017grad,springenberg2014striving,petsiuk2018rise,zintgraf2017visualizing,bach2015pixel,ribeiro2016should}) from attention-based methods (\textit{Attention}, e.g.~\cite{ba2014multiple, xu2015show, Fu_2017_CVPR}) that are usually incorporated into the network architecture to enforce a network's attention during training.

\paragraph{Dermoscopic Feature Prediction \& Localization} These methods aim at predicting whether dermoscopic features, relevant to the classification of skin diseases, are present in the input image or their localization. This type of explainability is often sometimes approached by means of multitask learning~\cite{caruana1997multitask}, which is problematic. A more sophisticated method that identifies abstract concepts in a pre-trained classifier was proposed by Kim et al.~\cite{kim2018interpretability}.

\paragraph{Similarity Retrieval} Another way of making predictions more intelligible is by allowing the user to retrieve cases, that share relevant similarities according to the classifier. Content-based Image Retrieval (CBIR) systems are usually trained for the very purpose of finding a semantically meaningful representation to retrieve similar images~\cite{wan2014deep}. However, the notion of CBIR can be applied to models trained on the primary objective of classification as well~\cite{qayyum2017medical}. 
%Those methods are grouped as~\textit{Similarity Retrieval} methods.

\paragraph{Intervention} The last group of methods is characterized by actively intervening and improving a model's explanations. These \textit{Intervention} methods include e.g. the penalization of wrong explanations during the training process~\cite{rieger2019interpretations} or identification of biases in the data as well as methods to remove them~\cite{lapuschkin2019unmasking}. 
%These methods are hereinafter referred to as methods of \textit{Intervention}. 
The continuous improvement of AI-based DSS in the health sector during their development and life cycle is crucial for their successful application~\cite{larson2020regulatory} and is therefore considered as a complement to explanation methods.
%between three methods that simply visualize network's activations\cite{}One class of methods that obtains such heatmaps leverages the network's gradients (e.g. Input$\times$Gradient~\cite{shrikumar2016not}, Class Activation Maps (CAM)~\cite{zhou2016learning}, GradCAM~\cite{selvaraju2017grad}, Guided Backpropagation~\cite{springenberg2014striving}). Perturbation-based methods perturb a region on the network's input to measure the impact on the predicted output (e.g. RISE~\cite{petsiuk2018rise}, Prediction Difference Analysis~\cite{zintgraf2017visualizing}). Decomposition-based methods like Layerwise Relevance Propagation (LRP)~\cite{bach2015pixel} decompose the prediction score into single input features to quantify their importance. The explanation of complex classifiers has also been attempted by locally approximating the decision boundary in methods like LIME~\cite{ribeiro2016should}. LIME produces spatial heatmaps as well. 
%Another way of revealing important input regions is to explicitly force a networks attention to informative region (e.g. Recurrent Attention CNN (RA-CNN)~\cite{Fu_2017_CVPR}). 
%Concept-based explanation methods like TCAV~\cite{kim2018interpretability} provide approximate global explanations by quantifying the influence of human-understandable concepts towards a class prediction. This also allows to validate the similarity of human-defined concepts with representations used by the DL-classifier.

\section{Overview of Methods}
\label{sec:overview}

Tables~\ref{tab:dermo} and~\ref{tab:histo} provide a comprehensive summary of all reviewed methods, including informations on the dataset, additional annotations and methods used.

\begin{table}[]
\caption{Clinical images and Dermoscopy: Comparison of explanatory approaches for clincial and dermoscopic image classifiers. Any non-public dataset used has been marked as private. Additional manual labeling and the use of quantitative measures of explainability are marked in separate columns.}
\label{tab:dermo}
    \resizebox{\textwidth}{!}{%
        \begin{tabular}{p{4.5cm}|c|c|c|c|c|c|c|c|c|c|c|c|c|c|p{.35cm}|p{.35cm}|c|c|p{6cm}}
        \noalign{\hrule height 1.5pt}
        
        \textbf{Author} & \multicolumn{1}{c|}{} & \multicolumn{2}{c|}{\textbf{Type}} &  \multicolumn{5}{c|}{\textbf{Datasets}} & \multicolumn{1}{c|}{} & \multicolumn{7}{c|}{\textbf{Method}} & \multicolumn{1}{c|}{} & \multicolumn{1}{c|}{} & \multicolumn{1}{c}{\textbf{Comment}}
        \\ %\hline
        \multicolumn{1}{c|}{} & \multicolumn{1}{c|}{} & \multicolumn{2}{c|}{} & \multicolumn{5}{c|}{} & \multicolumn{1}{c|}{} & \multicolumn{3}{c|}{Visual} & \multicolumn{1}{c|}{} & \multicolumn{2}{c|}{Feature} & \multicolumn{1}{c|}{} & \multicolumn{1}{c|}{} & \multicolumn{1}{c|}{} & \multicolumn{1}{c}{} 
        \\ %\hline
        \multicolumn{1}{c|}{} & \rotatebox[origin=c]{90}{\textbf{Year}} & \rotatebox[origin=c]{90}{Clinical} & \rotatebox[origin=c]{90}{Dermoscopic} & \rotatebox[origin=c]{90}{ISIC Archive} & \rotatebox[origin=c]{90}{Dermofit} & \rotatebox[origin=c]{90}{Derm7pt} & \rotatebox[origin=c]{90}{PH\textsuperscript{2}} & \rotatebox[origin=c]{90}{Private} & \rotatebox[origin=c]{90}{\textbf{Added Annotations}} & \rotatebox[origin=c]{90}{Activation} & \rotatebox[origin=c]{90}{Attribution} & \rotatebox[origin=c]{90}{Attention} & \rotatebox[origin=c]{90}{Similarity} & \rotatebox[origin=c]{90}{Prediction}  & \rotatebox[origin=c]{90}{Localization}  &  \rotatebox[origin=c]{90}{Intervention}  &  \rotatebox[origin=c]{90}{\textbf{Quantification}} &  \rotatebox[origin=c]{90}{\textbf{Evaluation}}  
        & \multicolumn{1}{c}{}
        \\ \noalign{\hrule height 0.5pt}
        
        \textbf{Radhakrishnan et al.~\cite{radhakrishnan2017patchnet}} & 17 & -- &\checkmark& \checkmark & -- & -- & -- & -- & -- & -- & \checkmark & -- & -- & -- & -- & -- & -- & \checkmark & Pseudo-Importance; Coarse patch-wise heatmaps.  \\ [1ex] \cline{1-20}
        \textbf{Esteva et al.~\cite{esteva2017dermatologist}} & 17 & \checkmark & \checkmark & \checkmark & \checkmark & -- & -- & \checkmark & -- & -- & \checkmark & -- & -- & -- & -- & -- & -- & -- & Basic attribution method (Saliency). \\ [1ex] \cline{1-20}
        \textbf{Ge et al.~\cite{ge2017skin}} & 17 & \checkmark & \checkmark & -- & -- & -- & -- & \checkmark & -- & -- & \checkmark & -- & -- & -- & -- & -- & -- & -- & Coarse heatmaps (CAM-BP). \\ [1ex] \cline{1-20}
        \textbf{Jia et al.~\cite{jia2017skin}} & 17 & -- & \checkmark & \checkmark & -- & -- & -- & -- & -- & -- & \checkmark & \checkmark & -- & -- & -- & -- & -- & -- & Coarse heatmaps (CAM). \\ [1ex] \cline{1-20}
        \textbf{Kawahara et al.~\cite{kawahara2018fully}} & 18 & -- & \checkmark & \checkmark & -- & -- & -- & -- & -- & -- & -- & -- & -- & -- & \checkmark & -- & -- & \checkmark & Dermoscopic feature segmentation. \\ [1ex] \cline{1-20}
        \textbf{Kawahara et al.~\cite{kawahara2018seven}} & 18 & \checkmark & \checkmark & -- & -- & \checkmark & -- & -- & -- & -- & -- & -- & \checkmark & \checkmark & \checkmark & -- & -- & -- & Use of metadata. Multi-task. \\ [1ex] \cline{1-20}
        \textbf{Li et al.~\cite{li2018evidence}} & 18 & -- & \checkmark & \checkmark & -- & -- & -- & -- & -- & -- & \checkmark & -- & -- & -- & -- & -- & -- & -- & Computationally intensive attribution. \\ [1ex] \cline{1-20}
        \textbf{Codella et al.~\cite{codella2018collaborative}} & 18 & -- & \checkmark & \checkmark & -- & -- & -- & -- & -- & -- & \checkmark & -- & \checkmark & \checkmark & -- & -- & -- & -- & Coarse heatmaps (QAM \& RAM). \\ [1ex] \cline{1-20}
        \textbf{Thandiackal et al.~\cite{thandiackal2018structure}} & 18 & -- & \checkmark & \checkmark & -- & -- & -- & -- & -- & -- & \checkmark & -- & -- & -- & \checkmark & -- & \checkmark & \checkmark & Multi-task. \\ [1ex] \cline{1-20}
        \textbf{Van Molle et al.~\cite{van2018visualizing}} & 18 & -- & \checkmark & \checkmark & -- & -- & -- & -- & -- & \checkmark & -- & -- & -- & -- & -- & -- & -- & -- &  \\ [1ex] \cline{1-20}
        \textbf{Yang et al.~\cite{yang2018classification}} & 18 & -- & \checkmark & \checkmark & -- & -- & -- & -- & -- & -- & \checkmark & -- & -- & -- & -- & -- & -- & -- & Coarse heatmap (CAM); Multi-task. \\ [1ex] \cline{1-20}
        \textbf{Yan et al.~\cite{yan2019melanoma}} & 19 & -- & \checkmark & \checkmark & -- & -- & -- & -- & -- & -- & -- & \checkmark & -- & -- & -- & \checkmark & -- & \checkmark & \\ [1ex] \cline{1-20}
        \textbf{Zhang et al.~\cite{zhang2019biomarker}} & 19 & -- & \checkmark & \checkmark & -- & -- & -- & -- & -- & -- & -- & -- & -- & -- & \checkmark & -- & -- & -- & Unsupervised biomarker localization; No distinction between biomarkers. \\ [1ex] \cline{1-20}
        \textbf{Mishra et al.~\cite{mishra2019interpreting}} & 19 & -- & \checkmark & -- & -- & -- & -- & \checkmark & -- & -- & \checkmark & -- & -- & -- & -- & -- & -- & -- & Coarse heatmaps (Grad-CAM, GBP). \\ [1ex] \cline{1-20}
        \textbf{Veltmeijer et al.~\cite{veltmeijer2019integrating}} & 19 & -- & \checkmark & \checkmark & -- & -- & -- & -- & \checkmark & -- & -- & -- & -- & \checkmark & \checkmark & -- & -- & -- & Multi-task.\\ [1ex] \cline{1-20}
        \textbf{Murabayashi et al.~\cite{murabayashi2019towards}} & 19 & -- & \checkmark & \checkmark & -- & -- & -- & \checkmark & -- & -- & -- & -- & -- & \checkmark & -- & -- & -- & -- & Multi-task; Semi-supervised VAT. \\ [1ex] \cline{1-20}
        \textbf{Xiang et al.~\cite{xiang2019towards}} & 19 & -- & \checkmark & \checkmark & -- & -- & -- & -- & -- & -- & \checkmark & -- & -- & -- & -- & -- & -- & -- & LIME attribution. \\ [1ex] \cline{1-20}
        \textbf{Rieger et al.~\cite{rieger2019interpretations}} & 19 & -- & \checkmark & \checkmark & -- & -- & -- & -- & -- & -- & \checkmark & -- & -- & -- & -- & \checkmark & -- & -- & Coarse heatmap (Grad-CAM). \\ [1ex] \cline{1-20}
        \textbf{Zhang et al.~\cite{zhang2019attention}} & 19 & -- & \checkmark & \checkmark & -- & -- & -- & -- & -- & -- & -- & \checkmark & -- & -- & -- & -- & -- & -- & Multi-stage attention. \\ [1ex] \cline{1-20}
        \textbf{Young et al.~\cite{young2019deep}} & 19 & -- & \checkmark & \checkmark & -- & -- & -- & -- & -- & -- & \checkmark & -- & -- & -- & -- & -- & -- & -- & Grad-CAM; Kernel SHAP. \\ [1ex] \cline{1-20}
        \textbf{Miko\l{}ajczyk et al.~\cite{mikolajczyk2020global}} & 20 & -- & \checkmark & \checkmark & -- & -- & -- & -- & -- & -- & \checkmark & -- & -- & -- & -- & \checkmark & -- & -- & Global explanation. \\ [1ex] \cline{1-20}
        \textbf{Sonntag et al.~\cite{sonntag2020skincare}} & 20 & -- & \checkmark & \checkmark & -- & -- & -- & -- & -- & -- & \checkmark & -- & -- & -- & \checkmark & -- & -- & -- & Coarse heatmaps (Grad-CAM, RISE); Multi-task. \\ [1ex] \cline{1-20}
        \textbf{Barata et al.~\cite{barata2020explainable}} & 20 & -- & \checkmark & \checkmark & -- & -- & -- & -- & -- & \checkmark & -- & \checkmark & -- & -- & -- & -- & -- & -- & Hierarchical approach. \\ [1ex] \cline{1-20}
        \textbf{Lucieri et al.~\cite{lucieri2020interpretability}} & 20 & -- & \checkmark & \checkmark & -- & \checkmark & \checkmark & \checkmark & -- & -- & -- & -- & -- & \checkmark & -- & -- & \checkmark & -- & Feature influence analysis on unconstrained DLS; Global explanation. \\ [1ex] \cline{1-20}
        \textbf{Coppola et al.~\cite{coppola2020interpreting}} & 20 & -- & \checkmark & -- & -- & \checkmark & -- & -- & -- & -- & -- & -- & -- & \checkmark & -- & -- & -- & -- & Multi-task. \\ [1ex] \cline{1-20}
        \textbf{Chen et al.~\cite{chen2020concept}} & 20 & -- & \checkmark & \checkmark & -- & -- & -- & -- & -- & -- & -- & -- & -- & \checkmark & -- & \checkmark & \checkmark & -- & Allows for concept discovery. \\ [1ex] \cline{1-20}
        \multicolumn{1}{r|}{\textbf{$\Sigma$}} &  & \textbf{3} & \textbf{26} & \textbf{22} & \textbf{1} & \textbf{3} & \textbf{1} & \textbf{5} & \textbf{1} & \textbf{2} & \textbf{14} & \textbf{4} & \textbf{2} & \textbf{7} & \textbf{6} & \textbf{4} & \textbf{3} & \textbf{4} & \\ \noalign{\hrule height 1.0pt}                    
\end{tabular}}
\end{table}

\begin{table}[]
\caption{Histopathology: Comparison of explanatory approaches for skin tissue classifiers. Any non-public dataset used has been marked as private. Additional manual labeling on public data and the use of quantitative measures of explainability are noted in separate columns.}
\label{tab:histo}
    \resizebox{\textwidth}{!}{%
        \begin{tabular}{p{4.5cm}|c|c|c|c|c|c|c|c|c|p{.35cm}|p{.35cm}|c|c|c}
        \noalign{\hrule height 1.5pt}
        \multicolumn{1}{c|}{\textbf{Author}} & \multicolumn{1}{c|}{} & \multicolumn{3}{c|}{\textbf{Datasets}} & \multicolumn{1}{c|}{} & \multicolumn{7}{c|}{\textbf{Method}} & \multicolumn{1}{c|}{}  & \multicolumn{1}{c}{\textbf{Comments}} 
        \\ %\hline
        \multicolumn{1}{c|}{} & \multicolumn{1}{c|}{} & \multicolumn{3}{c|}{} & \multicolumn{1}{c|}{} & \multicolumn{3}{c|}{Visual} & \multicolumn{1}{c|}{} & \multicolumn{2}{c|}{Feature} & \multicolumn{1}{c|}{}  & \multicolumn{1}{c|}{} & \multicolumn{1}{c}{} 
        \\ %\hline
        \multicolumn{1}{c|}{} & \multicolumn{1}{c|}{\rotatebox[origin=c]{90}{\textbf{Year}}} & \multicolumn{1}{c|}{\rotatebox[origin=c]{90}{TCGA}} & \multicolumn{1}{c|}{\rotatebox[origin=c]{90}{histologyDS}} & \multicolumn{1}{c|}{\rotatebox[origin=c]{90}{Private}} & \multicolumn{1}{c|}{\rotatebox[origin=c]{90}{\textbf{Added Annotations}}} & \multicolumn{1}{c|}{\rotatebox[origin=c]{90}{Activation}} & \multicolumn{1}{c|}{\rotatebox[origin=c]{90}{Attribution}} & \multicolumn{1}{c|}{\rotatebox[origin=c]{90}{Attention}} & \multicolumn{1}{c|}{\rotatebox[origin=c]{90}{Similarity}} & \multicolumn{1}{c|}{\rotatebox[origin=c]{90}{Prediction}} & \multicolumn{1}{c|}{\rotatebox[origin=c]{90}{Localization}} & \multicolumn{1}{c|}{\rotatebox[origin=c]{90}{Intervention}} & \multicolumn{1}{c|}{\rotatebox[origin=c]{90}{\textbf{Quantification}}} & \multicolumn{1}{c}{}
        \\ [10ex] \noalign{\hrule height 0.5pt}
        \multicolumn{1}{l}{\textbf{Cruz-Roa et al.~\cite{cruz2013deep}}} & \multicolumn{1}{|c|}{13} & -- & \checkmark & \checkmark & -- & -- & \checkmark & -- & -- & -- & -- & -- & -- & Pseudo-Importance; Coarse, patch-wise heatmap. \\ \cline{1-15}
        \multicolumn{1}{l}{\textbf{Cruz-Roa et al.~\cite{cruz2014automatic}}} & \multicolumn{1}{|c|}{14} & -- & -- & \checkmark & -- & -- & \checkmark & -- & -- & -- & -- & -- & -- & Pseudo-Importance; Coarse, patch-wise heatmap. \\ \cline{1-15}
        \multicolumn{1}{l}{\textbf{Xie et al.~\cite{xie2019interpretable}}} & \multicolumn{1}{|c|}{19} & \checkmark & -- & \checkmark & -- & -- & \checkmark & -- & -- & -- & -- & -- & -- & Coarse heatmaps (CAM, Grad-CAM). \\ \cline{1-15}
        \multicolumn{1}{l}{\textbf{Hägele et al.~\cite{hagele2020resolving}}} & \multicolumn{1}{|c|}{20} & \checkmark & -- & -- & \checkmark & -- & \checkmark & -- & -- & -- & -- & -- & \checkmark & Fine-grained heatmaps (LRP). \\ \cline{1-15}
        \multicolumn{1}{r|}{\textbf{$\Sigma$}} & & \textbf{2} & \textbf{1} & \textbf{3} & \textbf{1} & \textbf{0} & \textbf{4} & \textbf{0} & \textbf{0} & \textbf{0} & \textbf{0} & \textbf{0} & \textbf{1} & \\ \noalign{\hrule height 1.0pt}
\end{tabular}}
\end{table}

%\textit{Visual Relevance Localization} includes all methods and applications that mainly leverage visual attribution or attention maps of any kind. All methods that aim at explicitly classifying or localizing features that comprise interim tasks of the final prediction are grouped as \textit{Dermoscopic Feature Prediction \& Localization}. \textit{Similarity Retrieval} describes methods that explain decisions by presenting similar cases to the user. Finally, \textit{Intervention} includes approaches that allow to improve models or their explanations by incorporating expert knowledge or by removing bias.  

\subsection{Visual Relevance Localization}

\subsubsection{Visual Activation} Various methods have been proposed for the visualization of a CNNs intermediate feature representations (e.g. ~\cite{zeiler2014visualizing}, ~\cite{simonyan2013deep}).
Van Molle et al.~\cite{van2018visualizing} visualized the activation maps of a custom CNN trained on the ISIC archive by rescaling and mapping them onto the input image. They discovered separate activation maps excited by surrounding skin, hair-like features, variations in lesion colour and lesion borders. However, no activation maps have been found, focusing on common dermoscopic criteria as defined by dermatologists.

\subsubsection{Visual Attribution}

\begin{figure}
\centering
\includegraphics[width=\textwidth]{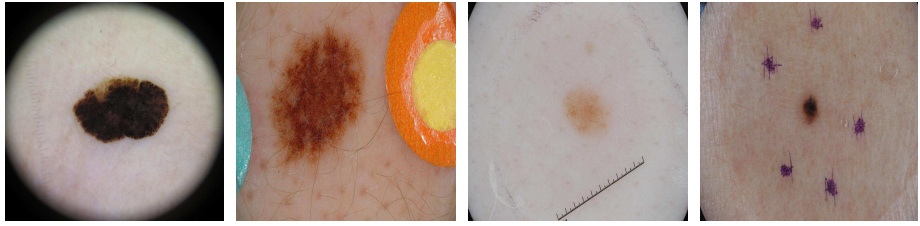}
\caption{Examples of skin images suffering from different artifacts like (from left to right) vignetting effects, coloured patches, scale, markings.}
\label{fig:example_artifacts}
\end{figure}

As early as 2013, Cruz-Roa et al.~\cite{cruz2013deep} worked on automated detection of BCC on the HistologyDS\footnote{Available at: http://www.informed.unal.edu.co/histologyDS} dataset using a patch-based DL approach. Their system processes WSIs patch-wise at a resolution of $8\times8$ through an encoder, a CNN and a final softmax classifier to get a prediction. A \textit{Digital Staining} procedure generates heatmaps on top of the input image by weighting each image patch with its resulting prediction score. In 2014, the authors proposed a revised method for detection of Invasive Ductal Carcinoma (IDC) on a private dataset in~\cite{cruz2014automatic}. Again, they applied a supervised, patch-base approach with a CNN to generate an IDC pseudo-probability map for visual analysis.
The 2017 work from Esteva et al.~\cite{esteva2017dermatologist} is well cited for being the first DL-based classifier outperforming 21 board-certified dermatologists in the classification of skin cancer. The authors used the GoogleNet InceptionV3~\cite{szegedy2016rethinking} architecture, pre-trained on ImageNet~\cite{russakovsky2015imagenet}, and fine-tuned on clinical and dermoscopic data from ISIC Archive, Dermofit and data from Stanford Hospital. Explanations are provided through saliency maps generated as gradients of the loss function w.r.t. the input. In the same year, Radhakrishnan et al.~\cite{radhakrishnan2017patchnet} proposed PatchNet, a patch-based architecture that processes each patch of a dermoscopic input image using a CNN. The patch-based paradigm is leveraged to compute attribution maps by combining the scores of every patch in the binary task of benign vs. malignant. Resulting attribution maps express the network's pseudo-probability of a patch being malignant. The patch-wise heatmaps are compared to Class Activation Maps (CAM)~\cite{zhou2016learning} and Grad-CAM~\cite{selvaraju2017grad} and quantitatively evaluated on segmentations of dermoscopic features provided in the ISIC 2017 challenge dataset.
%exhibiting a trade-off between heatmap granularity and global context along with generalization error. Their heatmaps outperformed CAM and GradCAM in matching dermoscopic features annotated as segmentation maps in the 2017 ISIC challenge dataset.
Ge et al.~\cite{ge2017skin} explored different multi-modal network structures, classifying pairs of clinical and dermoscopic images in one of 15 skin conditions. The private dataset used was provided by MoleMap\footnote{https://www.molemap.co.nz/}. Moreover, they propose CAM-BP which generates attribution maps by weighting Bilinear Pooling (BP)~\cite{lin2015bilinear} using CAM. In~\cite{li2018evidence}, the authors applied the perturbation-based \textit{Prediction Difference Analysis} method proposed in~\cite{zintgraf2017visualizing} to produce visual attribution maps.
%Therefore, they trained an ensemble of ResNet50 and VGG on the ISIC 2018 dataset.
In 2018, Yang et al.~\cite{yang2018classification} proposed \textit{Region Average Pooling} to improve dermoscopy image classification by jointly segmenting the lesion, using this segmentation map as ROI for pooling. CAM heatmaps were provided as well to prove the classifier's focus on the lesion region. 
In 2019, Xie et al.~\cite{xie2019interpretable} used VGG19~\cite{simonyan2014very} and ResNet50~\cite{he2016deep} networks to classify WSIs from the TCGA\footnote{Available at: https://portal.gdc.cancer.gov/} dataset into Melanoma and Nevi. CAM attribution has been computed to explain the classifiers.
Mishra et al.~\cite{mishra2019interpreting} trained several ResNet architectures and applied GradCAM~\cite{selvaraju2017grad} as well as Guided Backpropagation~\cite{springenberg2014striving} to analyze their models. The authors found that the models were having trouble with bad lighting, image blur, image noise and a wide field of view. Xiang et al.~\cite{xiang2019towards} trained an ensemble of different DL-based architectures. 
%AC-GAN was used to generate further samples for data augmentation. 
Explainability of their classifiers has been attempted by applying LIME~\cite{ribeiro2016should} to obtain visual attribution maps. It was confirmed that the classifiers made use of the lesion region for classification. 
%However, the limitations of the method have been pointed out as well.
Young et al.~\cite{young2019deep} trained 30 models on Melanoma and Nevi from a publicly available dermoscopic dataset and applied GradCAM and KernelSHAP~\cite{lundberg2017unified} for explanation. Despite high accuracies, models sometimes assigned relevance to features that were non-related to the actual classification task. Such artifacts can arise through process-specific conditions during clinical examination, bearing the risk of causing spurious correlations in the data. Well-known examples from public skin lesion datasets are the appearance of vignetting effects, coloured patches, scales or markings on skin lesion images as shown in Fig.~\ref{fig:example_artifacts}.
Recently, Hägele et al.~\cite{hagele2020resolving} emphasized on the importance of high-resolution attribution maps for the explanation and disclosure of biases in DL-based classifiers for histopathology. They compare high-resolution Layer-Wise Relevance Propagation (LRP)~\cite{bach2015pixel} maps to probability maps and GradCAM, demonstrating how fine-grained solutions allow explanations on cell-level instead of patch-level and how this is useful to discover biases in classifiers and data. To this end, they even provide a quantitative comparison of LRP and Grad-CAM by means of the area under the receiver operating characteristic curve (AUROC) for the detection sensitivity of cancer cells.

%Plenty of work has been published on explaining DL-based classifiers for histopathologic applications. However, the number of works specifically considering skin lesions is rather low. Three works are summarized hereinafter.

\subsubsection{Visual Attention} 
In 2017, Jia et al.~\cite{jia2017skin} developed a two-stage framework using a single CNN architecture that is first trained on ISIC 2017 dataset images. Then, Class Activation Maps are computed to crop images to the respective region of interest (ROI). The same CNN is then again trained with cropped images. The resulting classifier combines attribution and explicit attention for its explanation.
In the following year, Zhang et al.~\cite{zhang2019attention} proposed an \textit{Attention Residual Learning} (ARL) CNN that provides attention maps at every ARL block stage. Yan et al.~\cite{yan2019melanoma} incorporated an attention mechanism into a VGG16 network to obtain attention maps. In 2020, Barata et al.~\cite{barata2020explainable} treated the problem of skin lesion classification as a hierarchical problem of sequential word prediction. Using a CNN for feature encoding and a softmax attention mechanism for channel-wise and spatial attention masks, a masked embedding of the original image is fed into a recurrent long short-term memory (LSTM) module that sequentially predicts increasingly specific classes (Melanocytic vs. Non-Melanocytic; Benign vs. Malignant; Diagnosis). The attention maps were visualized, aiding the interpretability of the model.
%In addition to Barata et al.~\cite{barata2020explainable} used an MLP-based softmax attention mechanism for channel-wise and spatial attention masks.
Gu et al.~\cite{gu2020net} propose a comprehensive attention CNN (CA-Net) for medical image segmentation that combines the advantages of spatial, channel and scale attention to understand and interpret how pixel-level predictions are obtained.

\subsection{Dermoscopic Feature Prediction \& Localization}

%The segmentation and prediction of dermoscopic features is a line of work that facilitates the understanding and possible failure modes of classifiers, despite not being directly related to explainable AI.

In 2018, Kawahara et al.~\cite{kawahara2018fully} presented a fully convolutional architecture for the segmentation of four dermoscopic features annotated in the 2017 ISIC challenge dataset. Results are evaluated using a fuzzy Jaccard Index. In further work, Kawahara et al.~\cite{kawahara2018seven} developed a multi-modal DLS trained in a multi-task setting for the simultaneous detection and localization of diagnosis class and dermoscopic criteria. The CNN architecture is trained on sets of clinical and dermoscopic images along with meta data. It provides localization maps for all dermoscopic criteria, differianted in sub-types (e.g. regular / irregular streaks), can handle missing data (e.g. dermoscopic image without meta data) during inference due to the use of multiple loss combinations at train time and serves as a multi-modal image retrieval system. Thandiackal et al.~\cite{thandiackal2018structure} trained U-Nets~\cite{ronneberger2015u} to segment dermoscopic features. Those segmentation maps were stacked channel-wise with the original image and fed into a final CNN classifier 
%(ResNet50, ResNet18)
. Segmentation results are evaluated against the results from~\cite{kawahara2018fully}. Additionally, they allow to quantify the amount of network attribution (based on Input $\times$ Gradient) channel-wise and location-wise using partial derivatives. In 2019, Zhang et al.~\cite{zhang2019biomarker} attempted at localizing biomarkers without explicit labels. They trained a generative adversarial network (GAN)~\cite{goodfellow2014generative} on removing potential biomarkers from diseased images, to make them appear like benign images. Those fake benign images were subtracted from the GAN's input to localize potential dermoscopic criteria. The resulting representation was fed to a subsequent classifier to predict the initial image's class label. Veltmeijer et al.~\cite{veltmeijer2019integrating} manually annotated three dermoscopic criteria (Asymmetry, Irregular Borders and Multiple Colours) in the 2017 ISIC dataset. Using those annotations they trained two multi-task networks of which one predicted the criteria and the other segmented it, along with the diagnosis task. In~\cite{murabayashi2019towards}, Murabayashi et al. made use of semi-supervised \textit{Virtual Adversarial Training} (VAT) to leverage unlabeled data, training a multi-task CNN on the prediction of diagnosis labels and dermoscopic criteria. In 2020, Sonntag et al.~\cite{sonntag2020skincare} proposed a DSS for the explainable classification of skin lesions. The system allows lesion segmentation, dermoscopic criteria segmentation, binary or 8-class classification as well as the generation of visual attribution maps using GradCAM or RISE~\cite{petsiuk2018rise}. Lucieri et al.~\cite{lucieri2020interpretability} used TCAV~\cite{kim2018interpretability}, a concept-based explanation method, to find directions of dermoscopic features in the latent space of a trained CNN. Unlike all previously described methods, ~\cite{lucieri2020interpretability} analyse a trained CNN without constraining its training procedure by explicitly enforcing dermoscopic criteria learning. The authors found that a publicly available skin lesion classifier module, scoring third in the 2017 ISIC lesion classification challenge, is naturally able to distinguish between absence and presence of some common dermoscopic criteria defined by dermatologists. Furthermore, it was found that those criteria influence the CNNs decision in accordance with the medical consensus. Coppola et al.~\cite{coppola2020interpreting} recently proposed a novel multi-task learning network with learnable gates that allow to share features between subtasks. They discovered correlations between the features used for the detection of specific dermoscopic criteria and concluded that the diagnosis tasks uses features from most of the other subtasks, which confirms their intuition.

\subsection{Similarity Retrieval}

Codella et al.~\cite{codella2018collaborative} were the first to explicitly expand the explanation of dermoscopic classifiers by allowing content-based image retrieval (CBIR). By manually labeling images in non-expert similarity groups (e.g. visual appearance) and employing a triplet loss, their network's latent space is structured to provide meaningful nearest neighbours as evidence for its classification. In addition, they generate \textit{Result Activation Maps} (RAM) based on CAM, highlighting regions that contribute most to the proximity of the nearest neighbour of a sample. The previously mentioned multi-modal DLS by Kawahara et al.~\cite{kawahara2018seven} also allowed image retrieval. A special property is that a single feature vector can be used to retrieve multi-modal training inputs. 

\subsection{Intervention}

In 2019, Yan et al.~\cite{yan2019melanoma} improved the classification performance of their CNN by explicitly guiding the attention of the network through ground truth segmentation maps during training. Rieger et al.~\cite{rieger2019interpretations} introduced \textit{Contextual Decomposition Explanation Penalization} (CDEP). CDEP is a model-inherent method that allows to constrain the explanation of a deep network during training using ground-truth segmentation maps. As a result, bias can be removed iteratively during optimization, resulting in an explainable, accurate network. GradCAM is used to generate attribution maps. 
%(Explain in slightly more detail and maybe show image)
In 2020, Miko{\l}ajczyk et al.~\cite{mikolajczyk2020global} attempted global explanations by proposing \textit{Global Explanations for Discovering Bias in Data} (GEBI). The idea is based on \textit{SpRAy}~\cite{lapuschkin2019unmasking} which applies spectral clustering on a CNN's attribution maps to explain its behaviour. In~\cite{mikolajczyk2020global}, this method is extended by concatenating the input image to the corresponding attribution map before spectral clustering. Moreover, a counterfactual method is proposed in which a hypothesized bias is artificially induced in all images of the dataset to observe the change in predictive behaviour. Variations of LRP are used as attribution method. Building on the idea of TCAV, Chen et al.~\cite{chen2020concept} propose Concept Whitening (CW) as a normalization technique which is able to align a network's latent space with a set of desired concepts. Training a ResNet18 with CW on ISIC skin lesion images slightly improved the classification performance and resulted in a significantly more disentangled latent space. The authors concluded that the network did not focus on the age of a patient but instead on the size of the lesion. CW moreover reveals other concept directions that do not correspond to predefined concepts, simplifying the discovery of new, unknown biomarkers. The most influential concept axis in their experiments indicated the utilization of irregular borders for the detection of malignity.

% \section{Interesting Methods in Detail}

% Here I want to discuss at least Codella et al.~\cite{codella2018collaborative}, Rieger et al.~\cite{rieger2019interpretations} and Lucieri et al.~\cite{lucieri2020interpretability}. 

\section{Discussion}
\label{sec:discussion}
%The analysis of work on explainable deep skin lesion classifiers revealed current trends in the field.
Tables~\ref{tab:dermo} and~\ref{tab:histo} summarize all reviewed publications including additional information like datasets used, manual annotations, whether the approach allowed quantitative explanation and whether the explanations have been evaluated. It appears that most works aimed at explaining dermoscopic image classifiers. In spite of numerous publications on the explanation of DL-based classifiers for histopathology slides~\cite{zhang2017mdnet, yamamoto2019automated, sabol2020explainable, tosun2020histomapr}, only little attention has been drawn to skin tissue in particular. This might also be related to the limited availability of datasets and annotations, as is also reflected in the limited number of publicly available datasets used by the reviewed studies. 

It appears that two explanation modalities dominate previous work, namely visual attribution and dermoscopic criteria prediction \& localization. 
Visualization has always been a particularly popular way of explaining CNNs due to the importance of spatial arrangement in feature maps and visual input. However, the utility of different visualization methods varies a lot with their implementation. Numerous gradient-based attribution methods, for instance, have been shown to be independent of the model at hand~\cite{adebayo2018sanity} (e.g. Guided Backpropagation). Also, some methods produce coarse heatmaps owing to their implementation (e.g. CAM, GradCAM). Perturbation-based methods are known to suffer from out-of-distribution sampling which can significantly influence explanations~\cite{mittelstadt2019explaining}. Attention methods often explicitly enforce focus on specific image regions, thus limiting the information used by a DLS. Other approaches providing non-constraining attention cannot guarantee influence of their provided explanations on the prediction. The same holds for prediction and localization of dermoscopic criteria through multi-task optimization as well as simple activation visualization. 
%The visualization of network activations highlights regions that excite specific neurons. The observation of high activations is meaningless, as final predictions of DLS are complex non-linear combinations of all intermediate values. 
Some of the reviewed methods used the predicted scores of sub-regions (e.g. patches) to explain the network decision spatially. However, an opaque and uninterpretable high-level sub-decision can hardly be considered an explanation.

The implicit prediction and localization of dermoscopic criteria as practiced in~\cite{chen2020concept} and~\cite{lucieri2020interpretability} is closely related to the utilization of auxiliary classification objectives and hierarchization of classifier structures. These methods implicitly allow to monitor the DLS' understanding of predictions by verifying that a set of simpler sub-tasks are correctly performed and allow to quantify the influence of a sub-task to the final classifcation. The localization of single criteria is specifically useful to increase the efficiency of doctors by leading their attention and allowing them to validate network's decisions more easily. The Concept Localization Map (CLM) method proposed in~\cite{lucieri2020explaining} allows to extend frameworks based on CAVs for the implicit localization of biomarkers. Except for the method proposed in~\cite{zhang2019biomarker} and~\cite{chen2020concept}, all methods require the manual definition and annotation of criteria. This constrains the model performance in favour of explainability and constitutes a complex and laborious task, requiring hours of work from experts.

Attempts beyond visual and conceptual interpretability like content-based image retrieval approaches and methods that allow to explicitly adjust DLS's behaviour represent a significant step towards clinical usefulness of such systems. Some of the reviewed methods, for instance, allow to explicitly guide a network's focus and explanation through the incorporation of domain-knowledge or by detecting bias in the training data that can consequently be removed.

Analysis of existing work revealed that there is an ongoing trend of user-centered and adaptable explainable AI not only in the dermatology domain, but in AI in general. We suggest that future attempts to successfully deploy XAI in medicine should focus on following key points:
\begin{enumerate}
    \item User-centered explanations
    \item Diverse explanations
    \item Global explanations
    \item Interventions
\end{enumerate}
%Human-centered explanations refers to the utilization of concepts and ideas that are accessible to humans and common among domain experts.
\paragraph{User-Centered Explanations} The process of interpreting explanations is sometimes described as a translation of abstract concepts into a human-understandable domain~\cite{montavon2018methods}.
%Explaining in AI is often described as a transfer of knowledge that in AI aims to create understanding in the explainee about the decision that has been made by an agent. 
In order to create an understanding, it is essential to transfer knowledge in a way that is accessible to the human interpreter. 
%Currently and in the near future, explanations of AI will be directed towards humans. 
Thus, a user-centric orientation of explanations is a major requirement for effective explanations. The easiest way to fullfil this requirement is the communication by means of commonly used mental constructs (concepts) in the explainee's cultural milieu or the domain of expertise. This can, for instance, be implemented by explicitly handling intermediate tasks through modularization of architectures or the disclosure of such concepts in unconstrained DL-based algorithms. 
%This is important, as humans explaining their intentions or decisions usually rely on a foundation of common knowledge and conceptions. 

\paragraph{Diverse Explanations} Another crucial point that can be observed in the explanation of human beings and that will aid the interpretation of algorithmic decisions is the utilization of diverse modalities for explanation. Spatial localization of important input features using attribution maps is often helpful. However, only a fraction of the information necessary to explain a decision in its entirety is conveyed this way. The augmentation of visual relevance maps by more abstract, concept-based explanations and more fine-grained concept localization can result in an enriched and more complete explanation. Additionally, existing explanation approaches can be combined with different input modalities of the same data to reveal new relationships and increase diversity. Novel ideas considering input data in different data spaces such as frequency domain (e.g.~\cite{de2020human}), uncommon colour domains or other transformations (e.g.~\cite{abhishek2020illumination}) could potentially result in the alternative expression of relevance beyond spatial location. The combination of different explanation methods with diverse modalities on one hand allows for more complete explanations and on the other hand for the validation of the different explanations' congruence, hence the meaningfulness of the whole explanation.

\paragraph{Global Explanations} Most of today's xAI approaches consider only local explanations. However, a proper validation of network's working that will gain the trust of users and authorities can only be achieved by analyzing the bigger picture. In a DL setting, this means that all of the data space needs to be taken into account. So far, the consideration of the complete input space is - in most cases - impossible. Therefore, methods for the approximation of data spaces are required which in the best case provide some theoretical guarantees. As has been shown, current global explanations such as concept-based methods used in~\cite{lucieri2020interpretability} or the data-centric method proposed in~\cite{mikolajczyk2020global} approach global explanations by approximating the input space with as much data available. The combination of local and approximated global measures bears the potential to increase the scope of the explanation and therefore the user's understanding. 
%, as models are only as good as their training data. 
Without any guarantees of causality, assuring that a statistical model learnt only relevant coherences is only possible by identifying and removing systemic bias from the representation space. 
%Whether this is ultimately possible is another matter. 

\paragraph{Interventions} Ultimately, an explainable classifier is not worth much if it does not base its decision on legitimate grounds. Therefore, intervention is required to correct issues identified through explanations. Moreover, this is a way of explicitly incorporating expert knowledge into a learning system. The requirement for extensive annotation collection that is often seen as a drawback can be alleviated by applying classifiers with good base performance in clinical support settings, making use of interventional active learning. Here, the DLS represents the learner and an expert can act as a tutor that communicates potential misconceptions to the network on a case-by-case basis. The tutor benefits from the good and consistent performance of the learner, who in turn benefits form the correction of misconceptions. If proper channels of communication from tutor to student are provided, such scenarios might be comparable to a tutor teaching a real doctor in training.

Looking towards the future, another path should be followed in addition to the supervised incorporation of expert knowledge. As already mentioned, the definition and incorporation of human concepts and ideas into computer-understandable language is not trivial and requires exceptional efforts. Moreover, it is also known that many human decisions, processes and conventions emerged through time and are not always based on optimal solutions. Hence, it is not unlikely that there exist alternative representations of problems that, so far, seem abstract to humans but are well-grounded in theory. Decrypting the representations learned by high performing DL-based classifiers is therefore still of great potential value. First attempts towards unsupervised concept discovery have been proposed in~\cite{ghorbani2019towards, hu2020architecture, esser2020disentangling}. The most critical aspect is however the assignment of meaning exploitable by humans.

\section{Conclusions}
\label{sec:conclusion}
The present work reviewed and discussed the current state-of-the-art in explaining DL-based skin lesion classifiers. Four non-exclusive main groups of explanation approaches have been identified, namely, \textit{Visual Relevance Localization}, \textit{Dermoscopic Feature Prediction \& Localization}, \textit{Similarity Retrieval} as well as \textit{Intervention}. It has been shown that the explanation of histopathologic skin lesion classifiers has so far received little attention and that the main focus in the field lies on the explanation of dermoscopic classifiers by means of visual relevance maps and prediction \& localization of dermoscopic criteria. 
In order to meaningfully advance the field of medical AI towards practical, clinical deployment of DLS, we suggest to focus future efforts on user-centered explanation attempts that combine diverse modalities with local and global evidence. The continuous shaping of a classifier's behaviour through interventional active learning has the potential to revolutionize the way knowledge is transferred from human experts to DLS. Moreover, first attempts towards the unsupervised discovery of knowledge learned by data-driven algorithms indicate promising prospects for both, DL and the medical communities.

\bibliography{mybibfile}

\begin{thebibliography}{10}
\expandafter\ifx\csname url\endcsname\relax
  \def\url#1{\texttt{#1}}\fi
\expandafter\ifx\csname urlprefix\endcsname\relax\def\urlprefix{URL }\fi
\expandafter\ifx\csname href\endcsname\relax
  \def\href#1#2{#2} \def\path#1{#1}\fi

\bibitem{brinker2019deep}
T.~J. Brinker, A.~Hekler, A.~H. Enk, J.~Klode, A.~Hauschild, C.~Berking,
  B.~Schilling, S.~Haferkamp, D.~Schadendorf, T.~Holland-Letz, et~al., Deep
  learning outperformed 136 of 157 dermatologists in a head-to-head dermoscopic
  melanoma image classification task, European Journal of Cancer 113 (2019)
  47--54.

\bibitem{qiu2020development}
S.~Qiu, P.~S. Joshi, M.~I. Miller, C.~Xue, X.~Zhou, C.~Karjadi, G.~H. Chang,
  A.~S. Joshi, B.~Dwyer, S.~Zhu, et~al., Development and validation of an
  interpretable deep learning framework for alzheimer’s disease
  classification, Brain.

\bibitem{abramoff2018pivotal}
M.~D. Abr{\`a}moff, P.~T. Lavin, M.~Birch, N.~Shah, J.~C. Folk, Pivotal trial
  of an autonomous ai-based diagnostic system for detection of diabetic
  retinopathy in primary care offices, NPJ digital medicine 1~(1) (2018) 1--8.

\bibitem{PRNewswire2020}
P.~Newswire,
  \href{https://www.prnewswire.com/news-releases/3derm-announces-two-fda-breakthrough-device-designations-for-autonomous-skin-cancer-ai-300982072.html}{Fda
  permits marketing of artificial intelligence-based device to detect certain
  diabetes-related eye problems}, www.prnewswire.com.
\newline\urlprefix\url{https://www.prnewswire.com/news-releases/3derm-announces-two-fda-breakthrough-device-designations-for-autonomous-skin-cancer-ai-300982072.html}

\bibitem{FDA2018}
A.~Stark,
  \href{https://www.fda.gov/news-events/press-announcements/fda-permits-marketing-artificial-intelligence-based-device-detect-certain-diabetes-related-eye}{Fda
  permits marketing of artificial intelligence-based device to detect certain
  diabetes-related eye problems}, U.S. Food and Drug Administration (FDA).
\newline\urlprefix\url{https://www.fda.gov/news-events/press-announcements/fda-permits-marketing-artificial-intelligence-based-device-detect-certain-diabetes-related-eye}

\bibitem{rudin2019stop}
C.~Rudin, Stop explaining black box machine learning models for high stakes
  decisions and use interpretable models instead, Nature Machine Intelligence
  1~(5) (2019) 206--215.

\bibitem{szegedy2014intriguing}
C.~Szegedy, W.~Zaremba, I.~Sutskever, J.~Bruna, D.~Erhan, I.~Goodfellow,
  R.~Fergus, \href{http://arxiv.org/abs/1312.6199}{Intriguing properties of
  neural networks}, in: International Conference on Learning Representations,
  2014.
\newline\urlprefix\url{http://arxiv.org/abs/1312.6199}

\bibitem{bissoto2019constructing}
A.~Bissoto, M.~Fornaciali, E.~Valle, S.~Avila, (de) constructing bias on skin
  lesion datasets, in: Proceedings of the IEEE Conference on Computer Vision
  and Pattern Recognition Workshops, 2019, pp. 0--0.

\bibitem{european2020white}
E.~Commission, White paper on artificial intelligence - a european approach to
  excellence and trust.

\bibitem{schneeberger2020european}
D.~Schneeberger, K.~St{\"o}ger, A.~Holzinger, The european legal framework for
  medical ai, in: International Cross-Domain Conference for Machine Learning
  and Knowledge Extraction, Springer, 2020, pp. 209--226.

\bibitem{khazaei2019global}
Z.~Khazaei, F.~Ghorat, A.~Jarrahi, H.~Adineh, M.~Sohrabivafa, E.~Goodarzi,
  Global incidence and mortality of skin cancer by histological subtype and its
  relationship with the human development index (hdi); an ecology study in
  2018, World Cancer Research Journal (WCRJ) 6~(2) (2019) e13.

\bibitem{american20202cancer}
A.~C. Society, Cancer facts \& figures 2020, Am. Cancer Soc.

\bibitem{nachbar1994abcd}
F.~Nachbar, W.~Stolz, T.~Merkle, A.~B. Cognetta, T.~Vogt, M.~Landthaler,
  P.~Bilek, O.~Braun-Falco, G.~Plewig, The abcd rule of dermatoscopy: high
  prospective value in the diagnosis of doubtful melanocytic skin lesions,
  Journal of the American Academy of Dermatology 30~(4) (1994) 551--559.

\bibitem{argenziano1998epiluminescence}
G.~Argenziano, G.~Fabbrocini, P.~Carli, V.~De~Giorgi, E.~Sammarco, M.~Delfino,
  Epiluminescence microscopy for the diagnosis of doubtful melanocytic skin
  lesions: comparison of the abcd rule of dermatoscopy and a new 7-point
  checklist based on pattern analysis, Archives of dermatology 134~(12) (1998)
  1563--1570.

\bibitem{elmore2017pathologists}
J.~G. Elmore, R.~L. Barnhill, D.~E. Elder, G.~M. Longton, M.~S. Pepe, L.~M.
  Reisch, P.~A. Carney, L.~J. Titus, H.~D. Nelson, T.~Onega, et~al.,
  Pathologists’ diagnosis of invasive melanoma and melanocytic
  proliferations: observer accuracy and reproducibility study, Bmj 357 (2017)
  j2813.

\bibitem{arya2019one}
V.~Arya, R.~K. Bellamy, P.-Y. Chen, A.~Dhurandhar, M.~Hind, S.~C. Hoffman,
  S.~Houde, Q.~V. Liao, R.~Luss, A.~Mojsilovi{\'c}, et~al., One explanation
  does not fit all: A toolkit and taxonomy of ai explainability techniques,
  arXiv preprint arXiv:1909.03012.

\bibitem{arrieta2020explainable}
A.~B. Arrieta, N.~D{\'\i}az-Rodr{\'\i}guez, J.~Del~Ser, A.~Bennetot, S.~Tabik,
  A.~Barbado, S.~Garc{\'\i}a, S.~Gil-L{\'o}pez, D.~Molina, R.~Benjamins,
  et~al., Explainable artificial intelligence (xai): Concepts, taxonomies,
  opportunities and challenges toward responsible ai, Information Fusion 58
  (2020) 82--115.

\bibitem{van2018visualizing}
P.~Van~Molle, M.~De~Strooper, T.~Verbelen, B.~Vankeirsbilck, P.~Simoens,
  B.~Dhoedt, Visualizing convolutional neural networks to improve decision
  support for skin lesion classification, in: Understanding and Interpreting
  Machine Learning in Medical Image Computing Applications, Springer, 2018, pp.
  115--123.

\bibitem{barata2020explainable}
C.~Barata, M.~E. Celebi, J.~S. Marques, Explainable skin lesion diagnosis using
  taxonomies, Pattern Recognition (2020) 107413.

\bibitem{cruz2013deep}
A.~A. Cruz-Roa, J.~E.~A. Ovalle, A.~Madabhushi, F.~A.~G. Osorio, A deep
  learning architecture for image representation, visual interpretability and
  automated basal-cell carcinoma cancer detection, in: International Conference
  on Medical Image Computing and Computer-Assisted Intervention, Springer,
  2013, pp. 403--410.

\bibitem{cruz2014automatic}
A.~Cruz-Roa, A.~Basavanhally, F.~Gonz{\'a}lez, H.~Gilmore, M.~Feldman,
  S.~Ganesan, N.~Shih, J.~Tomaszewski, A.~Madabhushi, Automatic detection of
  invasive ductal carcinoma in whole slide images with convolutional neural
  networks, in: Medical Imaging 2014: Digital Pathology, Vol. 9041,
  International Society for Optics and Photonics, 2014, p. 904103.

\bibitem{radhakrishnan2017patchnet}
A.~Radhakrishnan, C.~Durham, A.~Soylemezoglu, C.~Uhler, Patchnet: Interpretable
  neural networks for image classification, arXiv preprint arXiv:1705.08078.

\bibitem{esteva2017dermatologist}
A.~Esteva, B.~Kuprel, R.~A. Novoa, J.~Ko, S.~M. Swetter, H.~M. Blau, S.~Thrun,
  Dermatologist-level classification of skin cancer with deep neural networks,
  nature 542~(7639) (2017) 115--118.

\bibitem{ge2017skin}
Z.~Ge, S.~Demyanov, R.~Chakravorty, A.~Bowling, R.~Garnavi, Skin disease
  recognition using deep saliency features and multimodal learning of
  dermoscopy and clinical images, in: International Conference on Medical Image
  Computing and Computer-Assisted Intervention, Springer, 2017, pp. 250--258.

\bibitem{jia2017skin}
X.~Jia, L.~Shen, Skin lesion classification using class activation map, arXiv
  preprint arXiv:1703.01053.

\bibitem{li2018evidence}
X.~Li, J.~Wu, E.~Z. Chen, H.~Jiang, What evidence does deep learning model use
  to classify skin lesions?, arXiv preprint arXiv:1811.01051.

\bibitem{codella2018collaborative}
N.~C. Codella, C.-C. Lin, A.~Halpern, M.~Hind, R.~Feris, J.~R. Smith,
  Collaborative human-ai (chai): Evidence-based interpretable melanoma
  classification in dermoscopic images, in: Understanding and Interpreting
  Machine Learning in Medical Image Computing Applications, Springer, 2018, pp.
  97--105.

\bibitem{thandiackal2018structure}
K.~Thandiackal, O.~Goksel, A structure-aware convolutional neural network for
  skin lesion classification, in: OR 2.0 Context-Aware Operating Theaters,
  Computer Assisted Robotic Endoscopy, Clinical Image-Based Procedures, and
  Skin Image Analysis, Springer, 2018, pp. 312--319.

\bibitem{yang2018classification}
J.~Yang, F.~Xie, H.~Fan, Z.~Jiang, J.~Liu, Classification for dermoscopy images
  using convolutional neural networks based on region average pooling, IEEE
  Access 6 (2018) 65130--65138.

\bibitem{mishra2019interpreting}
S.~Mishra, H.~Imaizumi, T.~Yamasaki, Interpreting fine-grained dermatological
  classification by deep learning, in: Proceedings of the IEEE Conference on
  Computer Vision and Pattern Recognition Workshops, 2019, pp. 0--0.

\bibitem{xiang2019towards}
A.~Xiang, F.~Wang, Towards interpretable skin lesion classification with deep
  learning models, in: AMIA Annual Symposium Proceedings, Vol. 2019, American
  Medical Informatics Association, 2019, p. 1246.

\bibitem{rieger2019interpretations}
L.~Rieger, C.~Singh, W.~J. Murdoch, B.~Yu, Interpretations are useful:
  penalizing explanations to align neural networks with prior knowledge, arXiv
  preprint arXiv:1909.13584.

\bibitem{young2019deep}
K.~Young, G.~Booth, B.~Simpson, R.~Dutton, S.~Shrapnel, Deep neural network or
  dermatologist?, in: Interpretability of Machine Intelligence in Medical Image
  Computing and Multimodal Learning for Clinical Decision Support, Springer,
  2019, pp. 48--55.

\bibitem{xie2019interpretable}
P.~Xie, K.~Zuo, Y.~Zhang, F.~Li, M.~Yin, K.~Lu, Interpretable classification
  from skin cancer histology slides using deep learning: A retrospective
  multicenter study, arXiv preprint arXiv:1904.06156.

\bibitem{mikolajczyk2020global}
A.~Miko{\l}ajczyk, M.~Grochowski, A.~Kwasigroch, Global explanations for
  discovering bias in data, arXiv preprint arXiv:2005.02269.

\bibitem{sonntag2020skincare}
D.~Sonntag, F.~Nunnari, H.-J. Profitlich, The skincare project, an interactive
  deep learning system for differential diagnosis of malignant skin lesions.
  technical report, arXiv preprint arXiv:2005.09448.

\bibitem{hagele2020resolving}
M.~H{\"a}gele, P.~Seegerer, S.~Lapuschkin, M.~Bockmayr, W.~Samek, F.~Klauschen,
  K.-R. M{\"u}ller, A.~Binder, Resolving challenges in deep learning-based
  analyses of histopathological images using explanation methods, Scientific
  reports 10~(1) (2020) 1--12.

\bibitem{yan2019melanoma}
Y.~Yan, J.~Kawahara, G.~Hamarneh, Melanoma recognition via visual attention,
  in: International Conference on Information Processing in Medical Imaging,
  Springer, 2019, pp. 793--804.

\bibitem{zhang2017mdnet}
Z.~Zhang, Y.~Xie, F.~Xing, M.~McGough, L.~Yang, Mdnet: A semantically and
  visually interpretable medical image diagnosis network, in: Proceedings of
  the IEEE conference on computer vision and pattern recognition, 2017, pp.
  6428--6436.

\bibitem{kawahara2018seven}
J.~Kawahara, S.~Daneshvar, G.~Argenziano, G.~Hamarneh, Seven-point checklist
  and skin lesion classification using multitask multimodal neural nets, IEEE
  journal of biomedical and health informatics 23~(2) (2018) 538--546.

\bibitem{kawahara2018fully}
J.~Kawahara, G.~Hamarneh, Fully convolutional neural networks to detect
  clinical dermoscopic features, IEEE journal of biomedical and health
  informatics 23~(2) (2018) 578--585.

\bibitem{veltmeijer2019integrating}
E.~Veltmeijer, S.~Karaoglu, T.~Gevers, et~al., Integrating clinically-relevant
  features into skin lesion classification., in: BNAIC/BENELEARN, 2019.

\bibitem{murabayashi2019towards}
S.~Murabayashi, H.~Iyatomi, Towards explainable melanoma diagnosis: Prediction
  of clinical indicators using semi-supervised and multi-task learning, in:
  2019 IEEE International Conference on Big Data (Big Data), IEEE, 2019, pp.
  4853--4857.

\bibitem{lucieri2020interpretability}
A.~Lucieri, M.~N. Bajwa, S.~A. Braun, M.~I. Malik, A.~Dengel, S.~Ahmed, On
  interpretability of deep learning based skin lesion classifiers using concept
  activation vectors, arXiv preprint arXiv:2005.02000.

\bibitem{coppola2020interpreting}
D.~Coppola, H.~Kuan~Lee, C.~Guan, Interpreting mechanisms of prediction for
  skin cancer diagnosis using multi-task learning, in: Proceedings of the
  IEEE/CVF Conference on Computer Vision and Pattern Recognition Workshops,
  2020, pp. 734--735.

\bibitem{chen2020concept}
Z.~Chen, Y.~Bei, C.~Rudin, Concept whitening for interpretable image
  recognition, Nature Machine Intelligence 2~(12) (2020) 772--782.

\bibitem{zhang2019biomarker}
R.~Zhang, S.~Tan, R.~Wang, S.~Manivannan, J.~Chen, H.~Lin, W.-S. Zheng,
  Biomarker localization by combining cnn classifier and generative adversarial
  network, in: International Conference on Medical Image Computing and
  Computer-Assisted Intervention, Springer, 2019, pp. 209--217.

\bibitem{zeiler2014visualizing}
M.~D. Zeiler, R.~Fergus, Visualizing and understanding convolutional networks,
  in: European conference on computer vision, Springer, 2014, pp. 818--833.

\bibitem{simonyan2013deep}
K.~Simonyan, A.~Vedaldi, A.~Zisserman, Deep inside convolutional networks:
  Visualising image classification models and saliency maps, arXiv preprint
  arXiv:1312.6034.

\bibitem{shrikumar2016not}
A.~Shrikumar, P.~Greenside, A.~Shcherbina, A.~Kundaje, Not just a black box:
  Learning important features through propagating activation differences, arXiv
  preprint arXiv:1605.01713.

\bibitem{zhou2016learning}
B.~Zhou, A.~Khosla, A.~Lapedriza, A.~Oliva, A.~Torralba, Learning deep features
  for discriminative localization, in: Proceedings of the IEEE conference on
  computer vision and pattern recognition, 2016, pp. 2921--2929.

\bibitem{selvaraju2017grad}
R.~R. Selvaraju, M.~Cogswell, A.~Das, R.~Vedantam, D.~Parikh, D.~Batra,
  Grad-cam: Visual explanations from deep networks via gradient-based
  localization, in: Proceedings of the IEEE international conference on
  computer vision, 2017, pp. 618--626.

\bibitem{springenberg2014striving}
J.~T. Springenberg, A.~Dosovitskiy, T.~Brox, M.~Riedmiller, Striving for
  simplicity: The all convolutional net, arXiv preprint arXiv:1412.6806.

\bibitem{petsiuk2018rise}
V.~Petsiuk, A.~Das, K.~Saenko, Rise: Randomized input sampling for explanation
  of black-box models, arXiv preprint arXiv:1806.07421.

\bibitem{zintgraf2017visualizing}
L.~M. Zintgraf, T.~S. Cohen, T.~Adel, M.~Welling, Visualizing deep neural
  network decisions: Prediction difference analysis, arXiv preprint
  arXiv:1702.04595.

\bibitem{bach2015pixel}
S.~Bach, A.~Binder, G.~Montavon, F.~Klauschen, K.-R. M{\"u}ller, W.~Samek, On
  pixel-wise explanations for non-linear classifier decisions by layer-wise
  relevance propagation, PloS one 10~(7) (2015) e0130140.

\bibitem{ribeiro2016should}
M.~T. Ribeiro, S.~Singh, C.~Guestrin, " why should i trust you?" explaining the
  predictions of any classifier, in: Proceedings of the 22nd ACM SIGKDD
  international conference on knowledge discovery and data mining, 2016, pp.
  1135--1144.

\bibitem{ba2014multiple}
J.~Ba, V.~Mnih, K.~Kavukcuoglu, Multiple object recognition with visual
  attention, arXiv preprint arXiv:1412.7755.

\bibitem{xu2015show}
K.~Xu, J.~Ba, R.~Kiros, K.~Cho, A.~Courville, R.~Salakhudinov, R.~Zemel,
  Y.~Bengio, Show, attend and tell: Neural image caption generation with visual
  attention, in: International conference on machine learning, 2015, pp.
  2048--2057.

\bibitem{Fu_2017_CVPR}
J.~Fu, H.~Zheng, T.~Mei, Look closer to see better: Recurrent attention
  convolutional neural network for fine-grained image recognition, in: The IEEE
  Conference on Computer Vision and Pattern Recognition (CVPR), 2017.

\bibitem{caruana1997multitask}
R.~Caruana, Multitask learning, Machine learning 28~(1) (1997) 41--75.

\bibitem{kim2018interpretability}
B.~Kim, M.~Wattenberg, J.~Gilmer, C.~Cai, J.~Wexler, F.~Viegas, et~al.,
  Interpretability beyond feature attribution: Quantitative testing with
  concept activation vectors (tcav), in: International conference on machine
  learning, 2018, pp. 2668--2677.

\bibitem{wan2014deep}
J.~Wan, D.~Wang, S.~C.~H. Hoi, P.~Wu, J.~Zhu, Y.~Zhang, J.~Li, Deep learning
  for content-based image retrieval: A comprehensive study, in: Proceedings of
  the 22nd ACM international conference on Multimedia, 2014, pp. 157--166.

\bibitem{qayyum2017medical}
A.~Qayyum, S.~M. Anwar, M.~Awais, M.~Majid, Medical image retrieval using deep
  convolutional neural network, Neurocomputing 266 (2017) 8--20.

\bibitem{lapuschkin2019unmasking}
S.~Lapuschkin, S.~W{\"a}ldchen, A.~Binder, G.~Montavon, W.~Samek, K.-R.
  M{\"u}ller, Unmasking clever hans predictors and assessing what machines
  really learn, Nature communications 10~(1) (2019) 1--8.

\bibitem{larson2020regulatory}
D.~B. Larson, H.~Harvey, D.~L. Rubin, N.~Irani, R.~T. Justin, C.~P. Langlotz,
  Regulatory frameworks for development and evaluation of artificial
  intelligence--based diagnostic imaging algorithms: Summary and
  recommendations, Journal of the American College of Radiology.

\bibitem{zhang2019attention}
J.~Zhang, Y.~Xie, Y.~Xia, C.~Shen, Attention residual learning for skin lesion
  classification, IEEE transactions on medical imaging 38~(9) (2019)
  2092--2103.

\bibitem{szegedy2016rethinking}
C.~Szegedy, V.~Vanhoucke, S.~Ioffe, J.~Shlens, Z.~Wojna, Rethinking the
  inception architecture for computer vision, in: Proceedings of the IEEE
  conference on computer vision and pattern recognition, 2016, pp. 2818--2826.

\bibitem{russakovsky2015imagenet}
O.~Russakovsky, J.~Deng, H.~Su, J.~Krause, S.~Satheesh, S.~Ma, Z.~Huang,
  A.~Karpathy, A.~Khosla, M.~Bernstein, et~al., Imagenet large scale visual
  recognition challenge, International journal of computer vision 115~(3)
  (2015) 211--252.

\bibitem{lin2015bilinear}
T.-Y. Lin, A.~RoyChowdhury, S.~Maji, Bilinear cnn models for fine-grained
  visual recognition, in: Proceedings of the IEEE international conference on
  computer vision, 2015, pp. 1449--1457.

\bibitem{simonyan2014very}
K.~Simonyan, A.~Zisserman, Very deep convolutional networks for large-scale
  image recognition, arXiv preprint arXiv:1409.1556.

\bibitem{he2016deep}
K.~He, X.~Zhang, S.~Ren, J.~Sun, Deep residual learning for image recognition,
  in: Proceedings of the IEEE conference on computer vision and pattern
  recognition, 2016, pp. 770--778.

\bibitem{lundberg2017unified}
S.~M. Lundberg, S.-I. Lee, A unified approach to interpreting model
  predictions, in: Advances in neural information processing systems, 2017, pp.
  4765--4774.

\bibitem{gu2020net}
R.~Gu, G.~Wang, T.~Song, R.~Huang, M.~Aertsen, J.~Deprest, S.~Ourselin,
  T.~Vercauteren, S.~Zhang, Ca-net: Comprehensive attention convolutional
  neural networks for explainable medical image segmentation, IEEE Transactions
  on Medical Imaging.

\bibitem{ronneberger2015u}
O.~Ronneberger, P.~Fischer, T.~Brox, U-net: Convolutional networks for
  biomedical image segmentation, in: International Conference on Medical image
  computing and computer-assisted intervention, Springer, 2015, pp. 234--241.

\bibitem{goodfellow2014generative}
I.~Goodfellow, J.~Pouget-Abadie, M.~Mirza, B.~Xu, D.~Warde-Farley, S.~Ozair,
  A.~Courville, Y.~Bengio, Generative adversarial nets, in: Advances in neural
  information processing systems, 2014, pp. 2672--2680.

\bibitem{yamamoto2019automated}
Y.~Yamamoto, T.~Tsuzuki, J.~Akatsuka, M.~Ueki, H.~Morikawa, Y.~Numata,
  T.~Takahara, T.~Tsuyuki, K.~Tsutsumi, R.~Nakazawa, et~al., Automated
  acquisition of explainable knowledge from unannotated histopathology images,
  Nature communications 10~(1) (2019) 1--9.

\bibitem{sabol2020explainable}
P.~Sabol, P.~Sin{\v{c}}{\'a}k, P.~Hartono, P.~Ko{\v{c}}an, Z.~Benetinov{\'a},
  A.~Blich{\'a}rov{\'a}, L.~Verb{\'o}ov{\'a}, E.~{\v{S}}tammov{\'a},
  A.~Sabolov{\'a}-Fabianov{\'a}, A.~Ja{\v{s}}kov{\'a}, Explainable classifier
  for improving the accountability in decision-making for colorectal cancer
  diagnosis from histopathological images, Journal of Biomedical Informatics
  109 (2020) 103523.

\bibitem{tosun2020histomapr}
A.~B. Tosun, F.~Pullara, M.~J. Becich, D.~L. Taylor, S.~C. Chennubhotla, J.~L.
  Fine, Histomapr™: An explainable ai (xai) platform for computational
  pathology solutions, in: Artificial Intelligence and Machine Learning for
  Digital Pathology, Springer, 2020, pp. 204--227.

\bibitem{adebayo2018sanity}
J.~Adebayo, J.~Gilmer, M.~Muelly, I.~Goodfellow, M.~Hardt, B.~Kim, Sanity
  checks for saliency maps, in: Advances in Neural Information Processing
  Systems, 2018, pp. 9505--9515.

\bibitem{mittelstadt2019explaining}
B.~Mittelstadt, C.~Russell, S.~Wachter, Explaining explanations in ai, in:
  Proceedings of the conference on fairness, accountability, and transparency,
  2019, pp. 279--288.

\bibitem{lucieri2020explaining}
A.~Lucieri, M.~N. Bajwa, A.~Dengel, S.~Ahmed, Explaining ai-based decision
  support systems using concept localization maps, in: International Conference
  on Neural Information Processing, Springer, 2020, pp. 185--193.

\bibitem{montavon2018methods}
G.~Montavon, W.~Samek, K.-R. M{\"u}ller, Methods for interpreting and
  understanding deep neural networks, Digital Signal Processing 73 (2018)
  1--15.

\bibitem{de2020human}
D.~de~Mijolla, C.~Frye, M.~Kunesch, J.~Mansir, I.~Feige, Human-interpretable
  model explainability on high-dimensional data, arXiv preprint
  arXiv:2010.07384.

\bibitem{abhishek2020illumination}
K.~Abhishek, G.~Hamarneh, M.~S. Drew, Illumination-based transformations
  improve skin lesion segmentation in dermoscopic images, in: Proceedings of
  the IEEE/CVF Conference on Computer Vision and Pattern Recognition Workshops,
  2020, pp. 728--729.

\bibitem{ghorbani2019towards}
A.~Ghorbani, J.~Wexler, J.~Y. Zou, B.~Kim, Towards automatic concept-based
  explanations, in: Advances in Neural Information Processing Systems, 2019,
  pp. 9277--9286.

\bibitem{hu2020architecture}
J.~Hu, R.~Ji, Q.~Ye, T.~Tong, S.~Zhang, K.~Li, F.~Huang, L.~Shao, Architecture
  disentanglement for deep neural networks, arXiv preprint arXiv:2003.13268.

\bibitem{esser2020disentangling}
P.~Esser, R.~Rombach, B.~Ommer, A disentangling invertible interpretation
  network for explaining latent representations, in: Proceedings of the
  IEEE/CVF Conference on Computer Vision and Pattern Recognition, 2020, pp.
  9223--9232.

\end{thebibliography}

\end{document}